\pdfoutput=1

\documentclass[11pt]{article}

\usepackage{ACL2023}
\usepackage{times}
\usepackage{latexsym}

\usepackage[T1]{fontenc}
\usepackage[utf8]{inputenc}
\usepackage{microtype}
\usepackage{inconsolata}
\usepackage{algorithm}
\usepackage{algorithmic}
\usepackage{makecell}
\usepackage{multirow}
\usepackage{CJKutf8}
\usepackage{color}
\usepackage{pgfplots}
\usepackage[normalem]{ulem}
\usepackage{graphicx}
\usepackage{amsfonts}
\usepackage{amsmath,amssymb,bm}
\usepackage{times}
\usepackage{latexsym}
\usepackage{color,xcolor}
\usepackage{CJKutf8}
\usepackage{amsmath,amsfonts}
\usepackage{algorithmic}
\usepackage{array}
\usepackage[caption=false,font=normalsize,
   labelfont=sf,textfont=sf]{subfig}
\usepackage{textcomp}
\usepackage{stfloats}
\usepackage{url}
\usepackage{verbatim}
\usepackage{graphicx}
\usepackage{balance}
\usepackage{inconsolata}
\usepackage{booktabs}
\usepackage{paralist}
\usepackage{colortbl}
\usepackage{algorithm}
\usepackage{algorithmic}
\usepackage{makecell}
\usepackage{arydshln}
\usepackage{multirow}
\usepackage{color}
\usepackage{pgfplots}
\usepackage[normalem]{ulem}
\usepackage{graphicx}
\usepackage{amsfonts}
\usepackage{amsmath,amssymb,bm}
\usepackage{soul}
\usepackage{amsmath}
\usepackage{tabularx}
\usepackage{booktabs}
\usepackage{amssymb}
\usepackage{pifont}
\usepackage{bm}
\usepackage{multirow}
\usepackage[framemethod=tikz]{mdframed}
\usepackage{tikz,pgfplots}
\usepgfplotslibrary{groupplots}
\pgfplotsset{compat=1.13}
\usepackage{pgf-pie}

\definecolor{remarkbg}{rgb}{0.927,1,1}
\definecolor{remarkborder}{gray}{0.15}
\colorlet{remarktitlebg}{remarkbg!20!black}
\usepackage[most]{tcolorbox}

\newenvironment{prompt_yellow}[1][]
  { 
 \begin{tcolorbox}
 [
    enhanced, 
    breakable,
    boxrule=0.5pt,
    arc=3.8pt,
    left=1.8pt,
    right=1.8pt,
    bottom=2pt,
    top=2pt, 
    rounded corners,
    colback=yellow!10
    ]{}
  \textbf{#1}
  \small \itshape
  }
  {
\end{tcolorbox} 
}

\definecolor{nmgray}{RGB}{229,229,229}
\definecolor{underlinegray}{RGB}{197,197,197}

\definecolor{introblue}{RGB}{0,176,240}
\definecolor{introgreen}{RGB}{0,203,134}
\definecolor{introgreen2}{RGB}{139,243,206}

\usepackage{adjustbox}

\newtcolorbox{mybox}[2][]{
width=\columnwidth,
colback = nmgray!75!white, 
colframe = nmgray!75!white, 
boxsep=0pt,left=10pt,right=10pt,top=0pt,bottom=0pt,
fontupper=\linespread{0.9}\selectfont,
title=#2,#1}

\usepackage{color}

\title{Dynamic Emotion and Personality Profiling for Multimodal Deception Detection}

\author{
  Li Zheng\textsuperscript{\rm 1},
  Yanyi Luo\textsuperscript{\rm 1}, Hao Fei\textsuperscript{\rm 2}, Yuzhe Ding\textsuperscript{\rm 1}, Yujie Huang\textsuperscript{\rm 1}, \\ \textbf{Fei Li\textsuperscript{\rm 1}\thanks{     
    $\,$ Corresponding author.}, Chong Teng\textsuperscript{\rm 1}, Donghong Ji\textsuperscript{\rm 1}\footnotemark[1]}
  \\
  \textsuperscript{\rm 1}Key Laboratory of Aerospace Information Security and Trusted Computing, Ministry of \\ Education, School of Cyber Science and Engineering, Wuhan University, Wuhan, China\\
  \textsuperscript{\rm 2}National University of Singapore, Singapore, Singapore
  \\
\texttt{\{zhengli,lne.luoyanyi,yuzheding,huang-yj\}@whu.edu.cn} \\ 
\texttt{\{lifei\_csnlp,tengchong,dhji\}@whu.edu.cn},  
    \texttt{haofei7419@gmail.com}}

\begin{document}
\begin{CJK}{UTF8}{gbsn}
\maketitle

\begin{abstract}\label{sec:abstract}
Deception detection is of great significance for ensuring information security and conducting public opinion analysis, with personality factors and emotion cues playing a critical role. 
However, existing methods lack sample-level dynamic annotations for emotions and personality.
In this paper, we propose an innovative multi-model multi-prompt annotation scheme and a strict label quality evaluation standard, and establish a multimodal joint detection dataset DDEP for deception, emotion, and personality. 
Meanwhile, we propose Rel-DDEP, an adaptive reliability-weighted fusion framework. 
Our framework quantifies uncertainty by mapping modal features to a high-dimensional Gaussian distribution space. It then performs reliability-weighted fusion and incorporates an alignment module and a sorting constraint module to achieve joint detection of deception, emotion, and personality.
Experimental results on the MDPE and DDEP datasets show that our Rel-DDEP significantly outperforms the existing state-of-the-art baseline models in three tasks. The F1 score of the deception detection increases by 2.53\%, that of the emotion detection increases by 2.66\%, and that of the personality detection increases by 9.30\%. 
The experiments fully verify the necessity of annotating dynamic emotion and personality labels for each sample and the effectiveness of reliability-weighted fusion. 
\end{abstract}

\section{Introduction}\label{sec:inroduction}

Deception detection aims to accurately identify deceptive behavior in individuals, which is critical for information security \cite{gutierrez2015modeling,han2018deception} and public opinion analysis \cite{alowibdi2014detecting,ding2025zero,zheng2025multi}. Prior works have yielded notable progress in multimodal deception detection. \citet{perez2015verbal} introduce a multimodal deception dataset using real-world courtroom trial videos. \citet{gupta2019bag} propose the Bag-of-Lies dataset for daily deceptive behavior detection.
\citet{vance2022deception} construct the multimodal deception database DDPM. 
Despite these advances, existing studies are confined to the single task of multimodal deception detection.

\begin{figure}[!t]
    \centering
    \includegraphics[scale=0.45]{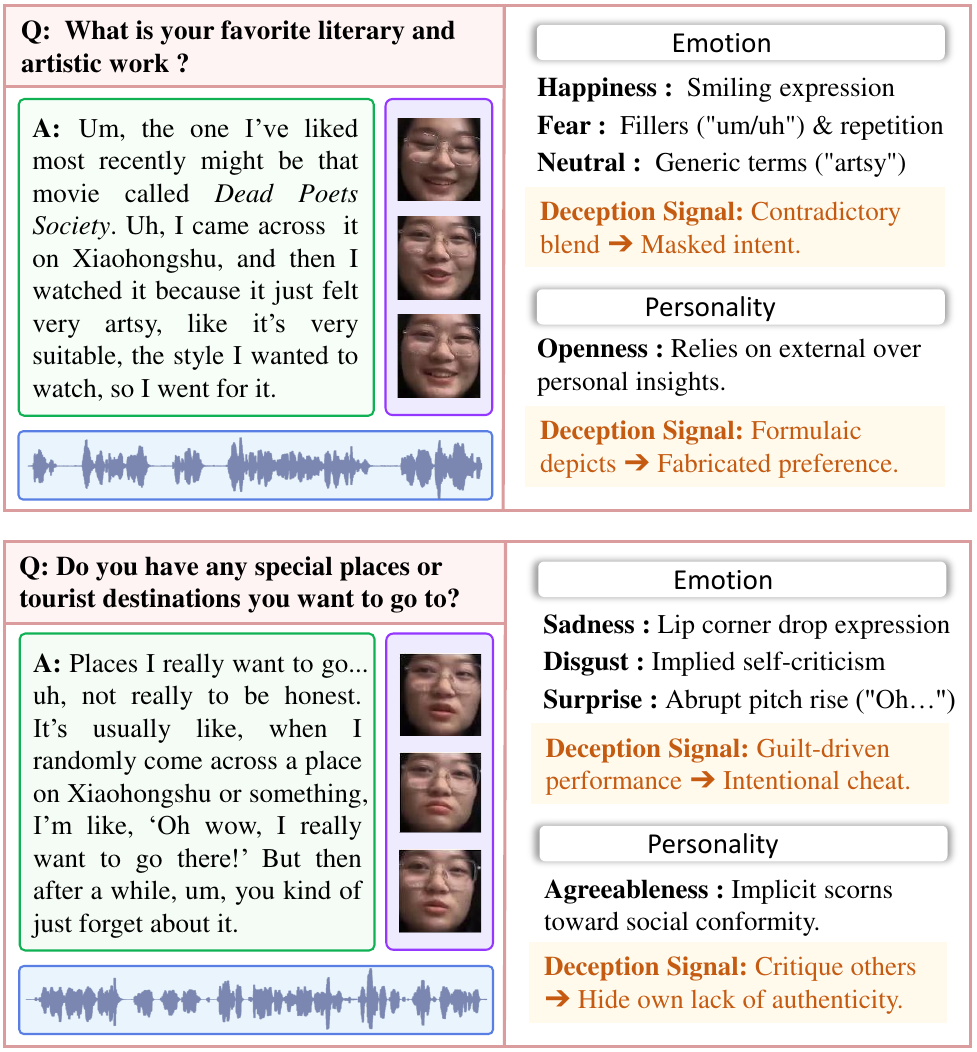}
    \vspace{-2mm}
    \caption{Examples of deceptive detection with different emotion and personality information.
    }
    \label{fig:ex}
    \vspace{-0.4cm}
\end{figure}

Numerous studies \cite{gaspar2013emotion,levitan2015cross,zheng2025improving} show that personality factors and emotion cues play important roles in deception detection. 
Personality factors affect an individual's behavior and ways of thinking, thus playing a role in deception detection. 
Emotion cues, whether they are genuine emotions or feigned emotions, can serve as key bases for detecting deception. 
Moreover, emotion and personality influence each other.
\citet{cai2024mdpe} propose a multimodal deception dataset with personality and emotion features.
However, their dataset only provides subject-level (per-participant) annotations, failing to capture sample-level dynamics, the fact that an individual’s emotions and personality can vary significantly across different situations.

As shown in Figure \ref{fig:ex}, the two examples correspond to the same subject. 
Subject-level annotation would assign identical emotion and personality features to both cases, yet their actual emotional and personality traits differ substantially. 
Furthermore, these discrepancies in emotion and personality provide critical cues for deception detection. 
The left example exhibits a blend of feigned happiness and fear of exposure, which is a hallmark of deceptive behavior. While the right example combines sadness and disgust, reflecting the subject’s internal disapproval of their perfunctory actions and facilitating deception identification as a result. 
Thus, sample-specific annotation is imperative.

Considering the complexity of emotions in real-world scenarios where emotions aren't presented in a single form, we adopt multi-label annotation for emotions (See the detailed analysis in Section \ref{Observation and Intuition}.). 
Since personality is relatively stable within a certain period, we set single-label annotation for personality.
Notably, manual annotation is costly, a longstanding major challenge in data annotation \cite{cai2021revisiting,hang2024dexfuncgrasp,zheng2024self}. 
Recent advances have shown that large language models (LLMs) possess strong text understanding and generation capabilities, and are increasingly used to assist human annotation tasks \cite{ding2022gpt,wang2024human}. 
However, the application of LLMs in multimodal annotation is still in the exploratory stage, with no unified standards for verifying annotation quality.

In view of this, we propose \textit{an innovative multi-model multi-prompt annotation scheme and a rigorous label quality evaluation standard}, and obtain the \textit{\underline{D}ynamic \underline{D}eception-\underline{E}motion-\underline{P}ersonality joint detection multimodal dataset \textbf{DDEP}}.
Specifically, we first employ multiple distinct LLMs for initial annotation and design diverse prompts for each model. 
These prompts guide LLMs to deeply understand data from varied perspectives, mitigating single perspective biases and reducing misjudgments. 
Subsequently, we design a voting mechanism and construct a quality scoring system incorporating consistency and uncertainty scores to comprehensively assess label quality. 
For data failing to meet the quality threshold, we leverage multimodal LLMs for advanced re-annotation followed by a second quality evaluation. 
Finally, data still not meeting requirements are re-annotated by professional human annotators.

We further propose \textit{\textbf{Rel-DDEP}, an adaptive reliability-weighted fusion framework for the joint multimodal detection of deception, emotion, and personality. }
Specifically, we project each modality’s features into a high-dimensional Gaussian distribution space to quantify uncertainty, then perform reliability-weighted fusion to assign rational weights to modalities (prioritizing highly reliable ones). 
Concurrently, we introduce an alignment module (to match uncertainty estimates with actual prediction errors) and a sorting constraint module (to ensure uncertainty estimates reflect modality importance order in joint detection). 
Finally, we derive the final predictions for emotion, personality, and deception.

\begin{figure*}[!h]
    \centering
    \includegraphics[scale=0.48]{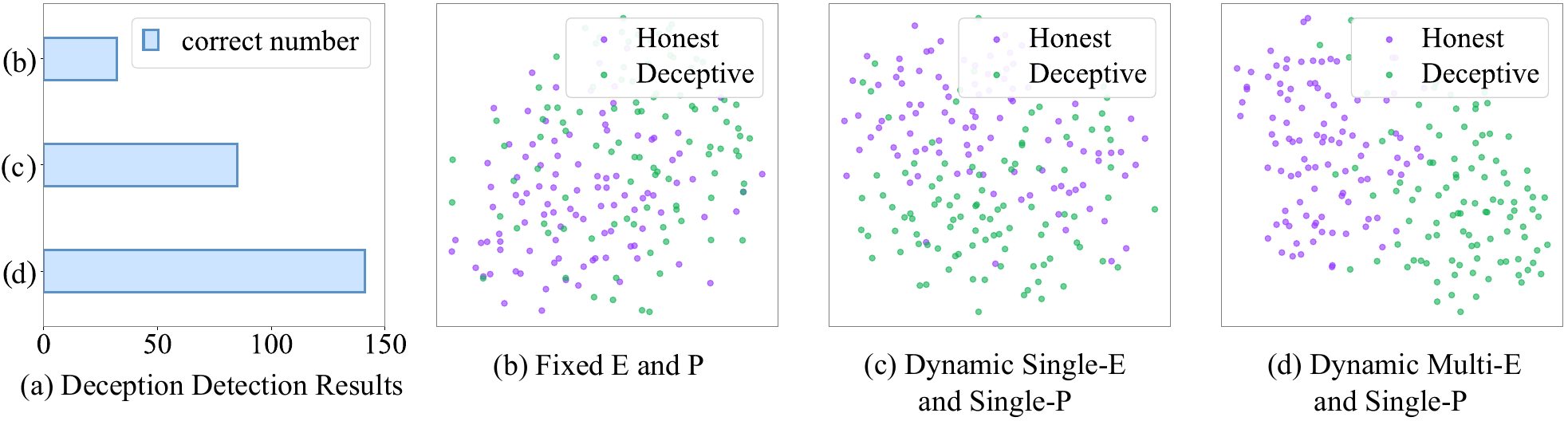}
    \vspace{-3mm}
    \caption{Visualization of different emotion (E) and personality (P) information.
    }
    \label{fig:tsne2}
    \vspace{-0.4cm}
\end{figure*}

To verify the effectiveness of our model, we conduct experiments on the widely used multimodal deception dataset MDPE \cite{cai2024mdpe} and our DDEP dataset.
The results show that our model significantly outperforms all state-of-the-art (SoTA) baselines on the three tasks of the two datasets. 
In the deception, emotion and personality detection task, the F1 score is increased by 2.53\%, 2.66\% and 9.30\% respectively.  
Extensive experiments verify the necessity of annotating dynamic emotion and personality labels for each data and the effectiveness of reliability-weighted fusion.

Our main contributions are summarized as follows:
\begin{itemize}
\item 

We propose a multi-model multi-prompt annotation scheme and a rigorous label quality evaluation standard, and establish a multimodal dataset DDEP for the joint detection of deception, emotion, and personality. 

\item 
We propose an adaptive reliability-weighted fusion framework to fully leverage the advantages of modalities with high reliability. 

\item 
Our extensive experimentation on the MDPE and DDEP datasets demonstrate that our Rel-DDEP achieves SoTA performance. 

\end{itemize}

\definecolor{color1}{HTML}{7FBF80}  
\definecolor{color2}{HTML}{E27386}  
\definecolor{color3}{HTML}{7EC8D1}  
\definecolor{color4}{HTML}{fe9778}  
\definecolor{color5}{HTML}{BFA0D3}  
\definecolor{color6}{HTML}{FFD66B}  
\definecolor{color7}{HTML}{D8BBAA}  
\definecolor{color8}{HTML}{A8A8A8}

\pgfplotsset{
    every axis/.append style={
        tick label style={font=\normalsize},
        label style={font=\normalsize},
        title style={font=\normalsize},
        legend style={font=\normalsize},
    }
}

\section{Observation and Key Intuition}
\label{Observation and Intuition}

Previous deception detection study \cite{cai2024mdpe} incorporates emotion and personality information at the subject-level. 
However, in reality, the emotions and personalities of the same person vary significantly in different situations.
To understand the impact of this variability on deception detection, we conduct a series of exploratory analyses.

We randomly select 20 subjects and choose 10 data instances with different emotion and personality characteristics for each subject. 
The deception detection results are shown in Figure \ref{fig:tsne2} (a). 
When relying solely on the fixed subject-level emotion and personality information (b) for deception detection, only 32 instances are correctly detected. 
This low success rate indicates that such fixed information is insufficient for effective deception detection. 
The visualization experiment in Figure \ref{fig:tsne2} (b) shows that deceptive and honest samples are intertwined in the visual space, with blurry boundaries that make them difficult to distinguish.
This visual evidence suggests that fixed subject-level annotations fail to capture the nuances that are crucial for accurate deception detection.

These findings inspire us to consider an alternative:\textit{ annotating each data instance with single emotion and personality labels. }
When applying this method to the same set of 200 samples, the number of correctly detected instances samples to 85, leading to an improvement in deception detection performance. 
The visualization results in Figure \ref{fig:tsne2} (c) also indicate that the distinction between deceptive and honest samples is enhanced.
This highlights the importance of annotating dynamic emotion and personality labels for individual data instances. 
However, there is still some overlap between deceptive and honest samples.

By observing the data, we find that the emotions in each sample are complex and diverse. 
For instance, the examples shown in Figure \ref{fig:ex} contain multiple emotions. 
This indicates that a more comprehensive annotation method is required. 
Therefore, \textit{we explore using multiple emotion labels for each sample while maintaining single-label annotations due to the relative stability of personality. }
Implementing this new strategy on the set of 200 samples brings about significant improvements, with 141 samples correctly detected. 
Figure \ref{fig:tsne2} (d) shows that deceptive and honest samples form distinct and compact clusters with concentrated feature distributions and clear boundaries. 
This result shows that dynamic multi-label emotion and single-label personality annotations can effectively capture sample-specific information, yielding more discriminative features for deception detection.
We formalize these findings into two theorems.

\noindent\textbf{Theorem 1.}  \textit{Information Gain Improvement Theorem.}
Let \(X_{fixed}\) be the feature set when using fixed labels, and \(X_{new}\) be the feature set after re-annotating the emotion and personality labels of each sample. Let \(Y\) be the category of deception detection results. Then, the information gain \(IG(Y|X_{new})>IG(Y|X_{fixed})\).

\noindent\textbf{Theorem 2.}  \textit{Situational Feature Difference Capture Theorem.}
Let the sample \(s\) be in different situations \(c_i\) and \(c_j\). With fixed labels, the feature difference degree \(D_{fixed}(X_{s,c_i},X_{s,c_j}) = 0\) (fixed labels ignore situational differences), and the degree after re-annotation is \(D_{new}(X_{s,c_i},X_{s,c_j})\). Then \(D_{new}(X_{s,c_i},X_{s,c_j})>0\).

\section{Multi-Model and Multi-Prompt Data Annotation}\label{sec:methodology}

\definecolor{color1}{HTML}{7FBF80}  
\definecolor{color2}{HTML}{E27386}  
\definecolor{color3}{HTML}{7EC8D1}  
\definecolor{color4}{HTML}{fe9778}  
\definecolor{color5}{HTML}{BFA0D3}  
\definecolor{color6}{HTML}{FFD66B}  
\definecolor{color7}{HTML}{D8BBAA}  
\definecolor{color8}{HTML}{A8A8A8}  

\pgfplotsset{
    every axis/.append style={
        tick label style={font=\normalsize},
        label style={font=\normalsize},
        title style={font=\normalsize},
        legend style={font=\normalsize},
    }
}

\begin{figure*}[!t]
    \centering
    \includegraphics[scale=0.48]{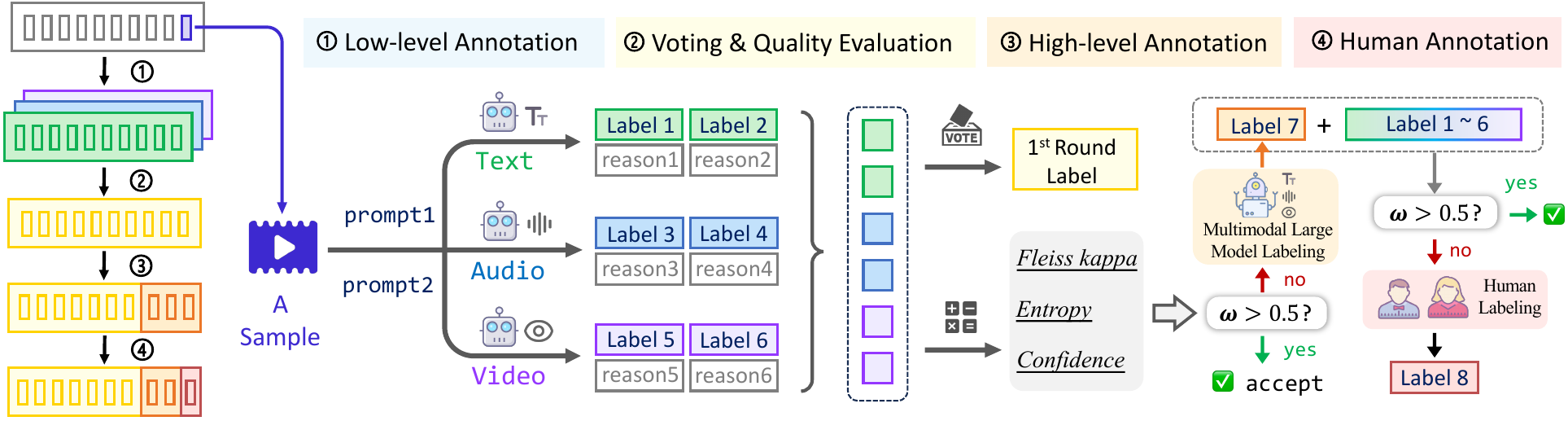}
    \caption{The flowchart of our data annotation. Label denote both emotion and personality. }
    \label{fig:data}
\end{figure*}

The MDPE dataset \cite{cai2024mdpe} offers rich data for multimodal deception detection, yet it suffers from limitations in emotion and personality annotation: it only provides subject-level emotion and personality labels, lacking sample-specific dynamic annotations. 
To better characterize data features per Theorems 1 and 2, we propose an innovative multi-model multi-prompt annotation scheme to assign emotion and personality labels to each sample, along with a set of label quality evaluation criteria. 
The annotation process is shown in Figure \ref{fig:data}.

\subsection{Low-level Annotation}

To comprehensively annotate the data, we select multiple LLMs of different types (e.g., GPT4o, Llama3, VideoLlama3, and Qwen2 Audio) for low-level annotation and design multiple prompting methods for each model. 
Through diverse prompts, we guide the models to understand the data from different perspectives, thereby generating more comprehensive annotation results. 
In the emotion annotation, one prompt guides the model to judge emotions from the overall atmosphere, as follows:

\begin{prompt_yellow} [Prompt1:]
\texttt{Please determine the emotions of the characters based on the overall atmosphere presented.}
\end{prompt_yellow}

Another prompt focuses on the emotions reflected by specific behaviors or expressions:

\begin{prompt_yellow} [Prompt2:]
\texttt{Please observe the facial expressions and body movements of the characters and determine their emotions.}
\end{prompt_yellow}

\subsection{Voting Mechanism and Annotation Quality Score Evaluation}

After generating annotations via the multi-model multi-prompt strategy, we derive the initial annotation round using a voting mechanism. 
For multi-label emotion annotation, we select labels with over half the votes as initial results. 
For personality labels, we choose the top-voted label if it garners more than half the votes.
If the number of votes for each label does not meet the requirement of more than half of the votes, it is recorded as an object that requires high-level re-annotation.
Our voting mechanism aggregates outputs from multiple models, effectively mitigating single-model errors and substantially boosting annotation reliability.

To ensure that the annotation quality meets a high standard, we construct a comprehensive quality score evaluation system, which is mainly composed of a consistency score and an uncertainty score.
The consistency score is measured by the kappa coefficient \cite{cohen1960coefficient}: 
\begin{equation}
k=\frac{p_{o}-p_{e}}{1 - p_{e}}
\end{equation}
where $p_{o}$ represents the observed consistency, 
$p_{e}$ represents the expected consistency. 
The value range of the Kappa coefficient is between -1 and 1.

The uncertainty score consists of two key metrics: entropy and self-evaluation confidence.
Entropy is used to measure the degree of uncertainty of a model's data classification. 
\begin{equation}
u_{i}=-\sum_{j = 1}^{n}p_{ij}\log(p_{ij})
\end{equation}
where \(p_{ij}\) denotes the probability that the \(i\)-th sample belongs to the \(j\)-th class, and \(n\) is the total number of classes. 
When the model is confident in a sample’s classification, the probability \(p_{ij}\) of the target class approaches 1 while those of other classes approach 0, yielding a low entropy value. 
Self-evaluation confidence refers to the model’s confidence in its annotation outputs. 
Specifically, we require each LLM to assign a confidence score \(s_c\) to its annotations with corresponding explanations. By jointly considering the consistency score and uncertainty score, we derive the final quality score:
\begin{equation}
S_q = \alpha_1 k+ \alpha_2u_i+\alpha_3s_c
\end{equation}

We screen the annotation results according to the set quality threshold. 
Annotations with a quality score lower than the threshold are considered to have low reliability and are marked as objects that need to be re-annotated at a high-level.

\begin{figure*}[htp!]
\centering
\resizebox{\textwidth}{!}{
\begin{tabular}{ccc}
\begin{tikzpicture}
\pie[
    explode={0.05},
    radius=2.5,
    color={color1!40,color5!40},
    draw=white, 
    sum=auto,
    after number = \%,
]{37.5/, 62.5/}

\node[anchor=west] at (3,1.2) {\LARGE \textcolor{color1!40}{\rule{0.7em}{0.7em}} \ Deceptive};
\node[anchor=west] at (3,0.45) {\LARGE \textcolor{color5!40}{\rule{0.7em}{0.7em}} \ Honest};

\node at (0,-3.5) {\huge Deception};
\end{tikzpicture}
&
\begin{tikzpicture}
\pie[
      explode={0.05,0.05,0.05,0.1,0.2,0.3,0.4,0.5},
      radius=2.5,
      after number = \%,
      draw=white,
      sum=auto,
      color={color1!40,color3!40,color5!40,color2!40,color4!40, color6!40, color7!40, color8!40}
    ]{%
      25.73/,
      24.95/,
      22.20/,
      9.38/,
      7.77/,
      5.77/,
      3.05/,
      0.69/
    }

    \node[anchor=west] at (3.3,2.2) {\LARGE \textcolor{color1!40}{\rule{0.7em}{0.7em}} \ Neutral};
    \node[anchor=west] at (3.3,1.45) {\LARGE \textcolor{color3!40}{\rule{0.7em}{0.7em}} \ Relaxation};
    \node[anchor=west] at (3.3, 0.7) {\LARGE \textcolor{color5!40}{\rule{0.7em}{0.7em}} \ Happiness};
    \node[anchor=west] at (3.3,-0.05) {\LARGE \textcolor{color2!40}{\rule{0.7em}{0.7em}} \ Sadness};
    \node[anchor=west] at (3.3,-0.8) {\LARGE \textcolor{color4!40}{\rule{0.7em}{0.7em}} \ Fear};
    \node[anchor=west] at (3.3,-1.55) {\LARGE \textcolor{color6!40}{\rule{0.7em}{0.7em}} \ Anger};
    \node[anchor=west] at (3.3,-2.3) {\LARGE \textcolor{color7!40}{\rule{0.7em}{0.7em}} \ Disgust};
    \node[anchor=west] at (3.3,-3.05) {\LARGE \textcolor{color8!40}{\rule{0.7em}{0.7em}} \ Surprise};

    \node at (0,-3.5) {\huge Emotion};
\end{tikzpicture}
&
\begin{tikzpicture}
\pie[
    explode={0.05,0.05,0.1,0.2,0.3},
    radius=2.5,
    after number = \%,
    color={color1!40,color3!40,color5!40,color2!40,color6!40},
    draw=white,
    sum=auto
]{%
    61.14/, 22.96/, 10.21/, 3.61/, 2.09/}

\node[anchor=west] at (3.3,1.2) {\LARGE \textcolor{color1!40}{\rule{0.7em}{0.7em}} \ Openness};
\node[anchor=west] at (3.3,0.45) {\LARGE \textcolor{color3!40}{\rule{0.7em}{0.7em}} \ Conscientiousness};
\node[anchor=west] at (3.3,-0.3) {\LARGE \textcolor{color5!40}{\rule{0.7em}{0.7em}} \ Extraversion};
\node[anchor=west] at (3.3,-1.05) {\LARGE \textcolor{color2!40}{\rule{0.7em}{0.7em}} \ Neuroticism};
\node[anchor=west] at (3.3,-1.8) {\LARGE \textcolor{color6!40}{\rule{0.7em}{0.7em}} \ Agreeableness};

\node at (0,-3.5) {\huge Personality};
\end{tikzpicture}
\end{tabular}}
\caption{The data distribution of our DDEP dataset.}
\label{fig:num}
\end{figure*}
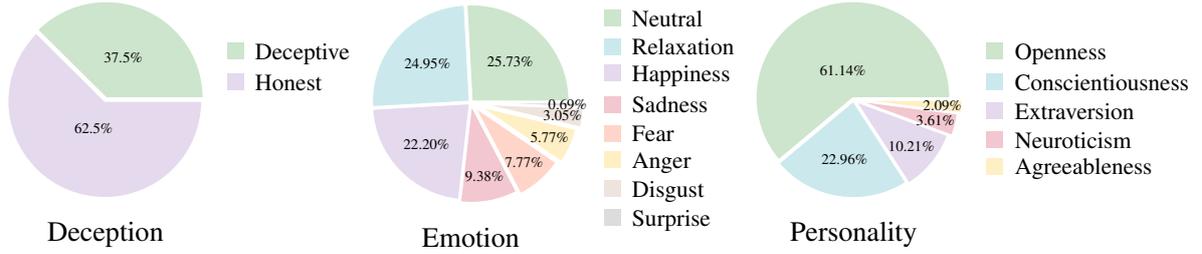

\begin{figure*}[!t]
    \centering
    \includegraphics[scale=0.48]{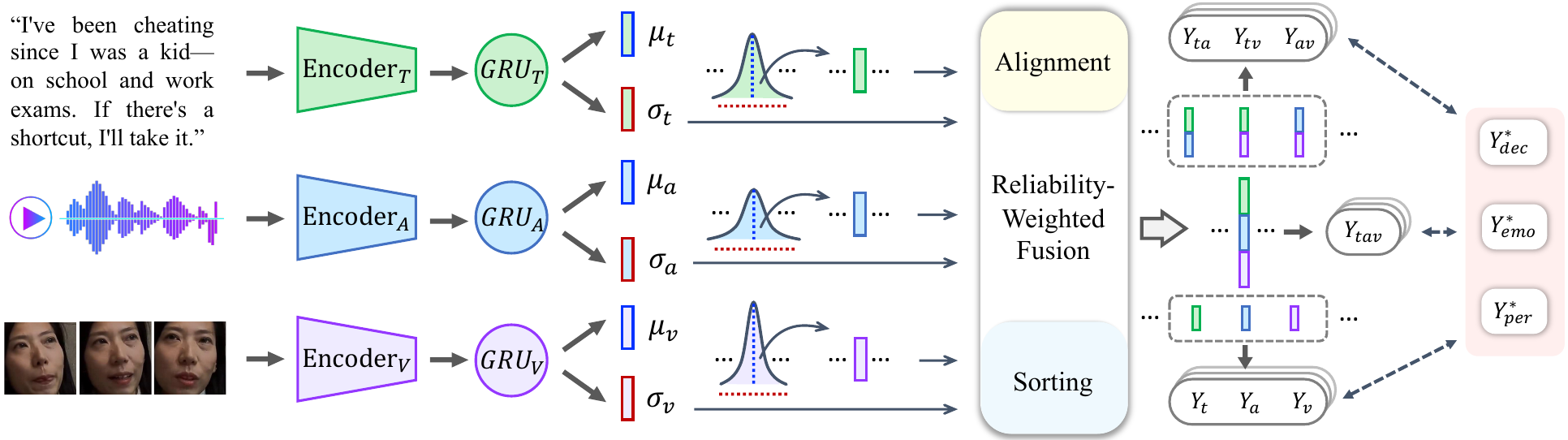}
    \caption{The overview of our framework.
    }
    \label{fig:model}
    \vspace{-0.4cm}
\end{figure*}

\subsection{High-level Annotation}

For the data marked as requiring high-level re-annotation, we use the multimodal large model as a high-level annotator to review the low-level annotation results of these unimodal LLMs and provide new annotation results along with explanations. 
This model conducts a more comprehensive and in-depth understanding and annotation of the data from the perspective of multimodal fusion.
Subsequently, we evaluate the quality of the annotation results to obtain the final quality scores.
If the quality score is lower than the threshold, the data is marked as requiring re-annotation and then handed over to human annotators for processing.

\subsection{Human Annotation}

Five natural language processing and emotion analysis experts conduct re-annotation.
During the re-annotation process, the annotators refer to the annotation results of LLMs. 
They repeatedly watch the videos and carefully read the texts, analyzing the emotion and personality information contained in the data from multiple perspectives to make comprehensive judgments.
To ensure the consistency and accuracy of the annotations, cross-validation is carried out on the annotation results of human annotators. 
Each data is independently annotated by at least two experts. 
Their results are then compared, and discrepancies are resolved through discussion until a consensus is reached. 
After annotation, we calculate the kappa score, achieving a score of \textbf{0.85}. 
We finally obtain the DDEP dataset, with labels illustrated in Figure \ref{fig:num}.

\section{Adaptive Reliability-weighted Fusion Network}

\subsection{Feature Extraction}

\noindent\textbf{Text extraction.} 
Text modal feature extraction aims to obtain key information such as semantics and context in the text. Following \citet{cai2024mdpe}, we adopt Baichuan (Bai) \cite{yang2023baichuan} for text feature extraction $\mathbf{h}_{t}$.

\noindent\textbf{Video extraction.} 
Following \citet{cai2024mdpe}, we use the multimodal models CLIP (CLB) \cite{radford2021learning} and ViT \cite{dosovitskiy2020image} to extract video features $\mathbf{h}_{v}$.

\noindent\textbf{Audio extraction.} 
Following \citet{cai2024mdpe}, we select Wav2vec-base (W2B) \cite{baevski2020wav2vec}, HUBERT-base (HBB) \cite{hsu2021hubert}, and WavLM-base (WMB) \cite{chen2022wavlm} to extract audio features $\mathbf{h}_{a}$.

\subsection{Uncertainty Estimation Module}

In the multimodal joint detection of deception, emotion, and personality, given the complexity of multimodal data and the inherent uncertainty of model predictions, accurately estimating uncertainty is of great importance.
For the features $\mathbf{h}_{m}$ ($m \in \{t, v, a\}$) extracted from each modality, we map them to a high-dimensional Gaussian distribution space $N(\mu_{m}, \sigma_{m})$ to quantify the uncertainty.
\begin{equation}
\mu_{m} = GRU_{\mu}(\mathbf{h}_{m}),
\sigma_{m} = GRU_{\sigma}(\mathbf{h}_{m})
\end{equation}
This uncertainty estimation module clarifies model prediction reliability, enabling rational weighting and fusion based on per-modality uncertainty.

\begin{table*}[!t]
\centering
\renewcommand{\arraystretch}{0.7} 
\setlength{\tabcolsep}{11pt} 
\begin{tabular}{llccccc} 
\toprule
\textbf{Task} & \textbf{Model} & \textbf{Acc.} & \textbf{AUC} & \textbf{P} & \textbf{R} & \textbf{F1} \\
\midrule
\multirow{9}{*}{Deception} 
& ViT-HBB-Bai-MDPE    & 64.00 & 67.50 & 47.15 & 52.91 & 49.86 \\
& ViT-HBB-Bai-DDEP    & 65.12 & 68.22 & 59.26 & 57.82 & 57.57 \\
& \cellcolor{gray!15}ViT-HBB-Bai-Ours & \cellcolor{gray!15}66.35 & \cellcolor{gray!15}69.45 & \cellcolor{gray!15}61.06 & \cellcolor{gray!15}59.41 & \cellcolor{gray!15}59.30 \\
& ViT-WMB-Bai-MDPE    & 63.59 & 67.20 & 46.89 & 54.35 & 50.35 \\
& ViT-WMB-Bai-DDEP    & 64.78 & 67.93 & 59.84 & 58.55 & 58.45 \\
& \cellcolor{gray!15}ViT-WMB-Bai-Ours & \cellcolor{gray!15}65.42 & \cellcolor{gray!15}69.15 & \cellcolor{gray!15}61.07 & \cellcolor{gray!15}60.10 & \cellcolor{gray!15}60.19 \\
& CLB-HBB-Bai-MDPE    & 64.66 & 68.70 & 49.48 & 52.33 & 50.87 \\
& CLB-HBB-Bai-DDEP    & 65.72 & 69.68 & 59.32 & 58.33 & 58.30 \\
& \cellcolor{gray!15}CLB-HBB-Bai-Ours & \cellcolor{gray!15}\textbf{66.63} & \cellcolor{gray!15}\textbf{70.21} & \cellcolor{gray!15}\textbf{61.49} & \cellcolor{gray!15}\textbf{60.71} & \cellcolor{gray!15}\textbf{60.83} \\
\midrule
\multirow{6}{*}{Emotion} 
& ViT-HBB-Bai         & 77.16 & 73.17 & 57.58 & 58.95 & 58.26 \\
& \cellcolor{gray!15}ViT-HBB-Bai-Ours & \cellcolor{gray!15}80.23 & \cellcolor{gray!15}79.02 & \cellcolor{gray!15}60.71 & \cellcolor{gray!15}60.43 & \cellcolor{gray!15}60.57 \\
& ViT-WMB-Bai         & 78.39 & 75.25 & 57.82 & 59.37 & 58.58 \\
& \cellcolor{gray!15}ViT-WMB-Bai-Ours & \cellcolor{gray!15}80.48 & \cellcolor{gray!15}78.51 & \cellcolor{gray!15}60.97 & \cellcolor{gray!15}61.26 & \cellcolor{gray!15}61.11 \\
& CLB-HBB-Bai         & 77.32 & 74.04 & 58.01 & 59.62 & 58.81 \\
& \cellcolor{gray!15}CLB-HBB-Bai-Ours & \cellcolor{gray!15}\textbf{80.73} & \cellcolor{gray!15}\textbf{79.49} & \cellcolor{gray!15}\textbf{61.16} & \cellcolor{gray!15}\textbf{61.78} & \cellcolor{gray!15}\textbf{61.47} \\
\midrule
\multirow{6}{*}{Personality} 
& ViT-HBB-Bai         & 82.51 & 72.11 & 39.51 & 38.22 & 38.98 \\
& \cellcolor{gray!15}ViT-HBB-Bai-Ours & \cellcolor{gray!15}85.55 & \cellcolor{gray!15}83.41 & \cellcolor{gray!15}51.01 & \cellcolor{gray!15}47.23 & \cellcolor{gray!15}49.15 \\
& ViT-WMB-Bai         & 83.89 & 74.91 & 40.78 & 39.89 & 40.30 \\
& \cellcolor{gray!15}ViT-WMB-Bai-Ours & \cellcolor{gray!15}85.58 & \cellcolor{gray!15}82.99 & \cellcolor{gray!15}51.63 & \cellcolor{gray!15}47.55 & \cellcolor{gray!15}49.71 \\
& CLB-HBB-Bai         & 83.07 & 75.02 & 40.92 & 40.33 & 40.60 \\
& \cellcolor{gray!15}CLB-HBB-Bai-Ours & \cellcolor{gray!15}\textbf{85.93} & \cellcolor{gray!15}\textbf{83.81} & \cellcolor{gray!15}\textbf{51.81} & \cellcolor{gray!15}\textbf{47.86} & \cellcolor{gray!15}\textbf{49.90} \\
\bottomrule
\end{tabular}
\caption{Experimental results on deception, emotion, and personality detection tasks.}
\label{tab:main}
\vspace{-0.4cm} 
\end{table*}

\subsection{Reliability-weighted Fusion Module}

Considering the differences in reliability of different modalities in the task, we introduce an reliability-weighted fusion module.
Based on the uncertainty estimation $\sigma_{m}$ of each modality, we calculate the fusion weight $w_{m}$. 
\begin{equation}
w_{m}=\frac{\frac{1}{\sigma_{m}}}{\sum_{j = t,v,a}\frac{1}{\sigma_{j}}}
\end{equation}
This formula ensures that the modality with lower uncertainty has a higher fusion weight. 
Then, we calculate the fused feature $\mathbf{h}_{f}$ by taking a weighted sum of each modality's features $\mathbf{h}_{m}$. 
\begin{equation}
\mathbf{h}_{f}=w_{t}\mathbf{h}_{t}+w_{v}\mathbf{h}_{v}+w_{a}\mathbf{h}_{a}
\end{equation}

\subsection{Alignment Module}
The alignment module aims to align the model’s uncertainty estimation with its actual prediction error. 
We define the alignment loss function $L_{ali}$, and use the mean squared error (MSE) to measure the difference between the uncertainty estimation $\sigma_{m}$ and the prediction error $\epsilon_{m}$. 
For each modality $m$, the prediction error $\epsilon_{m}$ is calculated by the cross-entropy loss between the predicted probability distribution $\mathbf{p}_{m}$ and the one-hot encoded distribution of the ground truth label $y$, that is:
\begin{equation}
\begin{split}
    \epsilon_{m} = CrossEntropyLoss(\mathbf{p}_{m}, \mathbf{y}_{one-hot}) \\
L_{ali}=\sum_{m = t,v,a}MSE(\sigma_{m}, \epsilon_{m})
\end{split}
\end{equation}

\subsection{Sorting Constraint Module}

The sorting constraint module enforces consistency between the uncertainty estimates of different modalities and the ranking of their fusion weights. 
We define the ordering loss $L_{\text{ord}}$, constructed by comparing the uncertainty estimates $\sigma_m$ and $\sigma_j$ of distinct modalities against the relative order of their corresponding fusion weights $w_m$ and $w_j$ ($m,j \in \{t,v,a\}$).
\begin{equation}
L_{sor}=\sum_{m \neq j}max(0, (\sigma_{m}-\sigma_{j}) - (\beta(w_{m}-w_{j})))
\end{equation}

\subsection{Prediction and Training}

The fused feature $\mathbf{h}_{f}$ passes through three fully connected layers $FC_{final}$ and a softmax function, obtaining final predictions $\hat{y}_d$, $\hat{y}_e$, $\hat{y}_p$ for deception, emotion, and personality tasks.
\begin{equation}
\hat{y}_{d/e/p} = softmax(FC_{final}(\mathbf{h}_{f}^{d/e/p}))
\end{equation}

We employ the cross-entropy loss function $L_{cls}$ to measure the difference between the final prediction result $\hat{y}_{d/e/p}$ and the ground-truth label $y_{d/e/p}$. 
By comprehensively considering the classification loss, alignment loss, and sorting loss, the overall loss function is formulated as:
\begin{equation}
L = L_{cls}+\lambda_{1}L_{ali}+\lambda_{2}L_{sor}
\end{equation}

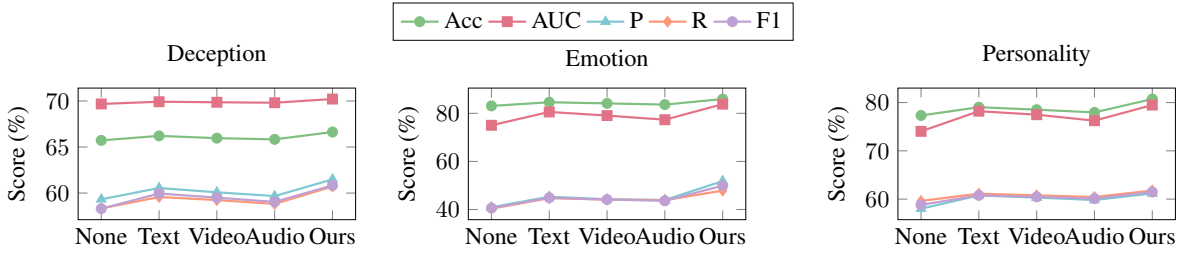
\begin{figure*}[htp!]
\centering
\resizebox{\textwidth}{!}{
\begin{tabular}{ccc}

\begin{tikzpicture}
\begin{axis}[
    title={Deception},
    height=0.48\columnwidth,
    width=0.78\columnwidth,
    ylabel={Score (\%)},
    xtick={0,1,2,3,4},
    xticklabels={None,Text,Video,Audio,Ours},
]

\addplot[mark=*, color=color1, line width=1pt] coordinates {(0,65.72) (1,66.21) (2,65.96) (3,65.83) (4,66.63)};
\addplot[mark=square*, color=color2, line width=1pt] coordinates {(0,69.68) (1,69.92) (2,69.86) (3,69.81) (4,70.21)};
\addplot[mark=triangle*, color=color3, line width=1pt] coordinates {(0,59.32) (1,60.55) (2,60.08) (3,59.67) (4,61.49)};
\addplot[mark=diamond*, color=color4, line width=1pt] coordinates {(0,58.33) (1,59.57) (2,59.24) (3,58.82) (4,60.71)};
\addplot[mark=*, color=color5, line width=1pt] coordinates {(0,58.30) (1,59.95) (2,59.50) (3,59.04) (4,60.83)};
\end{axis}
\end{tikzpicture} &

\begin{tikzpicture}
\begin{axis}[
    title={Emotion},
    height=0.48\columnwidth,
    width=0.78\columnwidth,
    ylabel={Score (\%)},
    xtick={0,1,2,3,4},
    xticklabels={None,Text,Video,Audio,Ours},
    legend style={at={(0.45,1.65)}, anchor=north, legend columns=-1}
]
\addplot[mark=*, color=color1, line width=1pt] coordinates { (0,83.07) (1,84.59) (2,84.13) (3,83.62)(4,85.93)};
\addplot[mark=square*, color=color2, line width=1pt] coordinates {(0,75.02) (1,80.57) (2,79.06) (3,77.33) (4,83.81)};
\addplot[mark=triangle*, color=color3, line width=1pt] coordinates {(0,40.92) (1,45.31) (2,44.27) (3,43.95) (4,51.81)};
\addplot[mark=diamond*, color=color4, line width=1pt] coordinates {(0,40.33) (1,44.72) (2,44.11) (3,43.89) (4,47.86)};
\addplot[mark=*, color=color5, line width=1pt] coordinates {(0,40.60) (1,44.90) (2,44.13) (3,43.62) (4,49.9)};
\legend{Acc, AUC, P, R, F1}
\end{axis}
\end{tikzpicture} &

\begin{tikzpicture}
\begin{axis}[
    title={Personality},
    height=0.48\columnwidth,
    width=0.78\columnwidth,
    ylabel={Score (\%)},
    xtick={0,1,2,3,4},
    xticklabels={None,Text,Video,Audio,Ours},
]
\addplot[mark=*, color=color1, line width=1pt] coordinates {(0,77.32) (1,79.04) (2,78.53) (3,77.96) (4,80.73)};
\addplot[mark=square*, color=color2, line width=1pt] coordinates {(0,74.04) (1,78.21) (2,77.49) (3,76.25) (4,79.49)};
\addplot[mark=triangle*, color=color3, line width=1pt] coordinates {(0,58.01) (1,60.76) (2,60.31) (3,59.82) (4,61.16)};
\addplot[mark=diamond*, color=color4, line width=1pt] coordinates {(0,59.62) (1,61.12) (2,60.78) (3,60.43) (4,61.78)};
\addplot[mark=*, color=color5, line width=1pt] coordinates {(0,58.81) (1,60.77) (2,60.46) (3,60.10) (4,61.47)};
\end{axis}
\end{tikzpicture} \\
\end{tabular}}
\caption{Influence of the reliability-weight fusion mechanism
on the DDEP dataset.
}
\label{fig:weight}
\vspace{-0.25cm}
\end{figure*}

\section{Experiments}\label{sec:experiments}
\subsection{Experimental Setup}\label{sec:experimental_setup}

\noindent\textbf{Dataset.} 
We evaluate the effectiveness of our model on the widely used MDPE dataset \cite{cai2024mdpe} and our newly established DDEP dataset.

\noindent\textbf{Baseline Systems.}
To validate the effectiveness of our model, we compare it with SoTA baselines tailored to each task.
\textbf{Deception Detection}: Two baseline categories are defined by dataset and feature strategy:
\textit{MDPE-based baselines}: Fixed emotion/personality feature models on the MDPE dataset (\texttt{-MDPE} in Table \ref{tab:main}).
\textit{DDEP-based baselines}: Dynamic emotion/personality feature models on our self-constructed DDEP dataset with only unimodal encoders (\texttt{-DDEP} in Table \ref{tab:main}).
\textbf{Emotion and Personality Detection}: All experiments are performed on the DDEP dataset (MDPE lacks corresponding labels), using competitive unimodal encoder-based models as baselines (details in Section 4.1).
For all tasks, \texttt{-Ours} in Table \ref{tab:main} denotes DDEP-evaluated models integrated with our reliability-weight fusion strategy.

\subsection{Main Results}

We conduct comprehensive experiments on the MDPE and DDEP datasets to systematically evaluate our method against SoTA baselines.
As shown in Table \ref{tab:main}, our method outperforms the SoTA baselines across all three tasks.
In the deception detection task, our method with dynamic emotion and personality labels far outperforms the fixed emotion and personality features at the subject-level. 
The F1 score increases by 2.53\%. 
This indicates that by annotating each sample with personalized emotion and personality labels, more clues related to deceptive behavior can be unearthed, thus significantly enhancing the performance of deception detection.
In emotion detection, our model shows a 2.66\% increase in the F1 score compared to the SoTA. 
In personality detection, the improvement of our model is even more remarkable. 
Our model has a 9.3\% increase in the F1 score compared to the SoTA baseline. 
This demonstrates that our reliability-weighted fusion can more rationally integrate multimodal data and effectively capture the key information related to emotions and personalities in different modalities.

\begin{table}[!t]
\centering
\small
\renewcommand{\arraystretch}{0.8} 
\setlength{\tabcolsep}{4pt} 
\begin{tabular}{l@{\hspace{6pt}}l@{\hspace{4pt}}ccccc} 
\toprule
\textbf{Task} & \textbf{Model} & \textbf{Acc} & \textbf{AUC} & \textbf{P} & \textbf{R} & \textbf{F1} \\
\midrule
\multirow{4}{*}{Dep} &  Ours  & \textbf{66.63}&\textbf{70.21}&\textbf{61.49}&\textbf{60.71}&\textbf{60.83}  \\
& \quad w/o Ali  & 66.32 & 70.04 & 60.89 & 60.24 & 60.58 \\
& \quad w/o Sort & 66.21 & 69.97 & 60.53 & 59.86 & 60.15 \\
& \quad w/o Rel  & 65.97 & 69.85 & 60.19 & 59.53 & 59.66 \\
\midrule
\multirow{4}{*}{Emo} &  Ours           & \textbf{80.73}&\textbf{79.49}&\textbf{61.16}&\textbf{61.78}&\textbf{61.47} \\
& \quad w/o Ali  & 79.68 & 77.92 & 60.41 & 61.12 & 60.59 \\
& \quad w/o Sort & 79.12 & 77.03 & 60.26 & 60.96 & 60.43 \\
& \quad w/o Rel  & 78.74 & 76.69 & 59.84 & 60.31 & 60.10 \\
\midrule
\multirow{4}{*}{Per} &  Ours & \textbf{85.93}&\textbf{83.81}&\textbf{51.81}&\textbf{47.86}&\textbf{49.90}  \\
& \quad w/o Ali  & 85.29 & 80.68 & 49.62 & 45.91 & 47.60 \\
& \quad w/o Sort & 84.81 & 79.25 & 48.92 & 45.28 & 47.11 \\
& \quad w/o Rel  & 84.47 & 78.31 & 46.79 & 43.84 & 45.21 \\
\bottomrule
\end{tabular}
\caption{Ablation study on multimodal deception detection task.}
\label{tab:abla}
\vspace{-0.4cm}
\end{table}

\subsection{Ablation Study}

To further explore the contributions of each component of our model, we conduct ablation experiments by removing alignment module, sorting constraint module, and reliability-weighted fusion module respectively.
The results are shown in Table \ref{tab:abla}.
Removing the alignment module causes consistent performance degradation across all metrics, underscoring its role in aligning uncertainty estimates with actual prediction errors. 
Similarly, eliminating the sorting constraint module reduces performance, demonstrating its necessity for ensuring modality uncertainty estimates reflect their importance in deception detection. 
Removing the reliability-weighted fusion module leads to the most significant decline, this highlights the module's ability to assign weights based on modality reliability, effectively emphasizing high reliability information.

\begin{figure*}[htp!]
\centering
\resizebox{\textwidth}{!}{ 
\begin{tabular}{ccc}
\begin{tikzpicture}
\begin{axis}[
    title={Deception},
    ybar,
    height=0.475\columnwidth,
    width=0.68\columnwidth,
    enlarge x limits=0.15,
    ylabel={Score (\%)},
    symbolic x coords={Acc,AUC,P,R,F1},
    xtick=data,
    ymin=50, ymax=75,
    bar width=7pt,
    legend style={at={(0.5,1.6)},anchor=north,legend columns=-1},
]
\addplot[draw=color3, fill=color3!30] coordinates {(Acc,64.32) (AUC,68.71) (P,59.28) (R,58.62) (F1,58.93)};
\addplot[draw=color5, fill=color5!30] coordinates {(Acc,66.63) (AUC,70.21) (P,61.49) (R,60.71) (F1,60.83)};
\end{axis}
\end{tikzpicture}&

\begin{tikzpicture}
\begin{axis}[
    title={Emotion},
    ybar,
    height=0.475\columnwidth,
    width=0.68\columnwidth,
    enlarge x limits=0.15,
    ylabel={Score (\%)},
    symbolic x coords={Acc,AUC,P,R,F1},
    xtick=data,
    ymin=40, ymax=85,
    bar width=7pt,
    legend style={at={(0.45,1.65)},anchor=north,legend columns=-1},
]
\addplot[draw=color3, fill=color3!30] coordinates {(Acc,77.45) (AUC,73.22) (P,58.46) (R,59.56) (F1,59.06)};
\addplot[draw=color5, fill=color5!30] coordinates {(Acc,80.73) (AUC,79.49) (P,61.16) (R,61.78) (F1,61.47)};
\legend{Single-task, Joint-task}
\end{axis}
\end{tikzpicture}&

\begin{tikzpicture}
\begin{axis}[
    title={Personality},
    ybar,
    height=0.475\columnwidth,
    width=0.68\columnwidth,
    enlarge x limits=0.15,
    ylabel={Score (\%)},
    symbolic x coords={Acc,AUC,P,R,F1},
    xtick=data,
    ymin=30, ymax=90,
    bar width=7pt,
]
\addplot[draw=color3, fill=color3!20] coordinates {(Acc,83.42) (AUC,76.32) (P,43.26) (R,42.17) (F1,42.82)};
\addplot[draw=color5, fill=color5!20] coordinates {(Acc,85.93) (AUC,83.81) (P,51.81) (R,47.86) (F1,49.90)};
\end{axis}
\end{tikzpicture} \\
\end{tabular}}
\caption{Comparative results of joint task and single task. }
\label{fig:joint}
\vspace{-0.3cm}
\end{figure*}
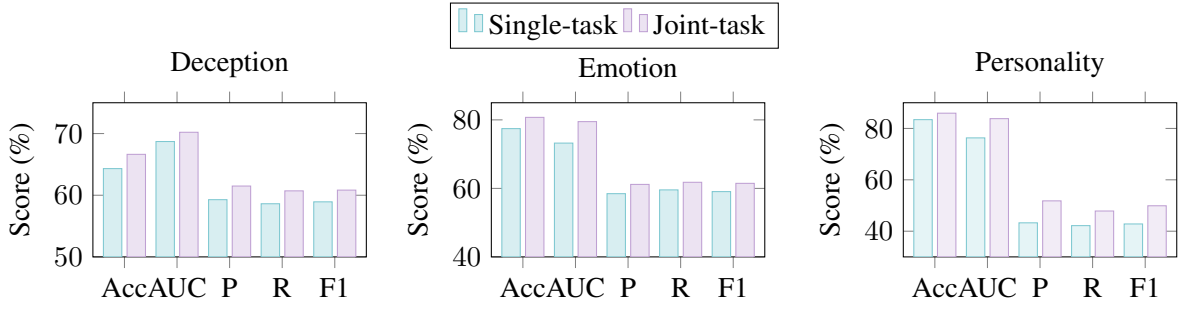

\subsection{The Impact of Reliability-Weighted Fusion Module}
To evaluate the impact of the reliability-weighted fusion module, Figure \ref{fig:weight} compares five strategies: reliability-weighted, no weights, and modality-dominant (text/audio/video) approaches. 
No weights fusion performs poorest, as equal treatment of modalities ignores their varying contributions, leading to information loss. 
Among modality-dominant baselines, text dominance yields the best results compared to audio dominant and video dominant approaches. 
This indicates that text plays a dominant role in the detection of these three tasks.
However, compared with the fusion approach using reliability weights, fixing a single modality as dominant still results in poor performance. 
Reliability-weighted dynamically assigns weights based on modality reliability, maximizing complementary strengths and enhancing robustness.

\subsection{The Effectiveness of Joint Tasks}
We design a set of single-task experiments for each task to thoroughly explore the effectiveness of the joint tasks. 
The experimental results are shown in Figure \ref{fig:joint}.
The results demonstrate that, in the three tasks of deception detection, emotion detection, and personality detection, the performance of the joint task significantly outperforms that of single tasks. This indicates that there are close inherent connections and synergistic effects among emotions, personality, and deception. 
In the joint-task mode, the information related to each task can complement and promote one another, thereby enhancing the overall performance.

\section{Related Work}\label{sec:related_works}

\noindent\textbf{Deception Detection.} 
Deceptive behaviors are widespread in numerous fields such as social interactions, business, and security \cite{damstra2021does,kumar2021deception,zheng2025ecqed,zheng2025enhancing}. 
\citet{abouelenien2016detecting} integrate multiple modalities of information to construct a comprehensive deception detection system. 
\citet{mathur2020introducing} focus on analyzing the discriminative abilities of visual, speech, and language modality features and explore their impact mechanisms on deception detection.
Regarding dataset construction, previous studies \cite{gupta2019bag,perez2015verbal,vance2022deception} have contributed a series of valuable deception detection datasets. 
Although these researches achievements are remarkable, they only focus on the single task of deception. 
Considering that personality factors and emotion cues play crucial roles in deception detection, \citet{cai2024mdpe} propose a multimodal deception dataset with personality and emotion features.

\noindent\textbf{LLMs Assisted Data Annotation.} 
The rapid development of machine learning highly depends on a large amount of labeled training data. 
Yet, annotation is costly and time-consuming. 
A promising solution combines automatic pre-annotation with human refinement.
\citet{wang2021want} re-annotate low-confidence LLM-generated instances, while \cite{gilardi2023chatgpt,he2023annollm,zheng2025stpar,zheng2024reverse} show ChatGPT outperforms humans in multiple tasks, validating LLMs’ task-specific strengths.
\citet{ding2022gpt} use GPT-3 as an annotator for classification and token-level task experiments.
\citet{wang2024human} propose a multi-stage human-LLM collaboration method to optimize annotation workflows. 
Prior works \cite{kneusel2017improving,lai2019human} further confirm that model outputs enhance human decision-making, highlighting LLMs’ annotation support value.

\section{Conclusion}
\label{sec:Conclusion}

In this paper, we propose an innovative method for joint detection of multimodal deception, emotion, and personality, which integrates dynamic emotion and personality annotation and an adaptive reliability-weighted fusion framework. 
By implementing a multi-model multi-prompt annotation strategy and  a label quality evaluation standard, we construct the high-quality DDEP dataset. 
Our Rel-DDEP quantifies uncertainties and conducts reliability-weighted fusion, effectively enhancing the performance of the model. 
Numerous experiments on the MDPE and our DDEP datasets show that our framework achieves state-of-the-art results.

\section*{Limitations}

Our work uses a basic shared encoder for the joint recognition of emotion, personality, and deception tasks. While this design ensures computational efficiency and simplifies training by reusing cross-task feature extraction layers, it inherently limits deep interactions among the three tasks. Specifically, the shared encoder learns general-purpose features uniformly applied to all tasks, failing to capture task-specific nuances and context-aware cross-task correlations. In future research, we will explore advanced multi-task learning frameworks to explicitly model task-specific dependencies, dynamically allocate feature extraction resources, and enhance adaptability to diverse task interactions, thereby further improving joint recognition performance.

\section*{Acknowledgments}

This work was funded by Kuaishou.
Fei Li and Donghong Ji are co-corresponding authors.

\bibliographystyle{acl_natbib}
\bibliography{main}

\end{CJK}
\end{document}